\documentclass[11pt,a4paper]{article}
\usepackage[hyperref]{ranlp2021}
\usepackage{times}
\usepackage{latexsym}

\usepackage{microtype}

\aclfinalcopy

\usepackage{url}
\usepackage{rotating}
\usepackage{amssymb}
\usepackage{graphicx}
\usepackage{multirow}
\usepackage{comment}
\usepackage{paralist,lipsum}
\usepackage{amsmath}
\usepackage{mathtools}
\usepackage[subrefformat=parens]{subcaption}

\usepackage{makecell}

\newcommand{\oldalign}{\textsc{FastAlign}} 
\newcommand{\newalign}{\textsc{BabAlign}} 
\newcommand{\sensembert}{\textsc{SensEmBERT}}
\newcommand{\unsup}{\textsc{LabelSync}}
\newcommand{\semisup}{\textsc{LabelProp}}
\newcommand{\labelgen}{\textsc{LabelGen}}
\newcommand{\softconstraint}{\textsc{SoftConstraint}}

\title{Semi-Supervised and Unsupervised Sense Annotation via Translations}
 
\author{
 Bradley Hauer,
 Grzegorz Kondrak,
 Yixing Luan,
 Arnob Mallik,
 Lili Mou\\
 Alberta Machine Intelligence Institute, Department of Computing Science\\
 University of Alberta, Edmonton, Canada\\
 {\tt {\{bmhauer,gkondrak,yixing1,amallik,lmou\}}@ualberta.ca}
}

\date{}

\begin{document}
\maketitle

\begin{abstract}
Acquisition of multilingual training data
continues to be a challenge in word sense disambiguation (WSD).
To address this problem,
unsupervised approaches have been proposed
to automatically generate sense annotations
for training supervised WSD systems.
We present three new methods for creating sense-annotated corpora
which leverage translations, parallel bitexts, lexical resources, 
as well as contextual and synset embeddings.
Our semi-supervised method applies machine translation
to transfer existing sense annotations to other languages.
Our two unsupervised methods 
refine sense annotations produced by a knowledge-based WSD system 
via lexical translations in a parallel corpus.
We obtain state-of-the-art results on standard WSD benchmarks.
\end{abstract}

\section{Introduction}
\label{sec:intro}

Word sense disambiguation,
the task of
identifying the meaning of a word in context,
is one of the central problems in
natural language understanding \cite{navigli2018}.
It is a well-studied benchmark for evaluating
contextualized representations of words 
\cite{loureiro2021},
and is better understood than tasks such as WiC \cite{pilehvar2019}.
Modern WSD methods can be divided into
supervised and knowledge-based approaches.
The former depend on {\rm sense-annotated corpora},
such as SemCor \cite{miller1994},
while
the latter
rely instead on semantic knowledge bases
such as WordNet \cite{miller1995acm}.

While
supervised WSD systems
typically outperform knowledge-based systems \cite{scarlini2020b},
their utility is limited by
the availability of
sufficiently large sense-annotated corpora 
for training.
This includes 
systems based on contextualized embeddings
\cite{bevilacqua2020}.
In particular, there is a severe lack of high-quality sense-annotated corpora
for languages other than English.
This
limitation
has motivated
the development of methods
aimed at
automatically disambiguating a large number of word tokens
in a given unannotated corpus,
ideally covering a wide range of word and sense types,
while minimizing noise
\cite{pasini2017, scarlini2019, barba2020}.
The automatically tagged corpus
can then be used to train a supervised WSD system,
satisfying the dependency on training data
without the need for manual annotation.

Following recent theoretical work \newcite{hauer2020set}
on establishing the semantic equivalence of mutual translations,
we introduce three translation-based methods 
for generating sense-tagged corpora.
All three methods make use of lexical knowledge bases,
and semantic information obtained from word-level translations.
Semi-supervised {\semisup} 
creates a synthetic parallel corpus ({\em bitext})
by applying machine translation to a monolingual manually-annotated corpus,
and projecting annotations to the target language.
Similarly,
unsupervised \labelgen{} 
applies a knowledge-based WSD system to the English side of a bitext,
and projects the resulting sense annotations across
bitexts onto other languages.
Finally,
unsupervised {\unsup} 
produces sense-annotated corpora in two languages at once
by independently applying a knowledge-based WSD system
to each side of a raw bitext,
and then refining 
the initial annotations
based on the confidence scores and multilingual information.

Our experiments on standard WSD test sets
demonstrate that the new methods achieve state-of-the-art results
in both semi-supervised and unsupervised
sense annotation.
We train
two different reference supervised WSD systems on the generated data,
and apply the resulting models to multilingual WSD benchmarks.
Our results compare favourably to
models trained on data produced by 
the previous state-of-the-art sense annotation methods.
Indeed, 
some of the results obtained with our unsupervised methods
rival those 
obtained by training on a manually sense-annotated corpus.

Our contributions are as follows:
We present three novel, 
scalable methods 
that can generate annotated corpora for any language
for which a suitable lexical knowledge base is available.
We show that these methods achieve state-of-the-art results
on multiple languages. 
We make our code and corpora 
available.\footnote{https://www.cs.ualberta.ca/$\sim${}kondrak}

\section{Related Work}
\label{sec:rw}

\begin{figure}[t]
  \centering
  \includegraphics[keepaspectratio, width=\columnwidth]
  {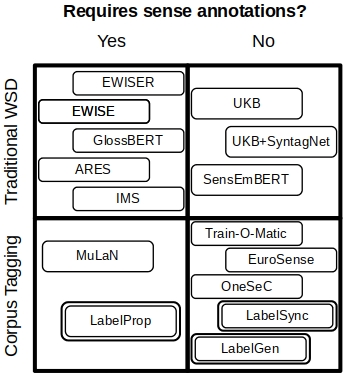}
 \caption{Typology of relevant sense tagging systems.
Our own systems are shown in 
double-lined boxes.}
 \label{fig_relwork}
\end{figure}

The sense tagging systems that we consider in this work,
including our three novel methods,
can be divided into four types according to two criteria
(Figure \ref{fig_relwork}).
The first criterion is whether the method involves supervision
in the form of a sense-annotated corpus.
The second criterion is whether the method
operates as a traditional self-contained WSD system,
or instead assigns
sense tags to a subset of the words in a corpus
which can then be used to train a supervised WSD system.
In this section, we discuss the most relevant examples
of each of the four resulting types.

{\bf Supervised WSD} systems rely on sense annotations
to train disambiguation models,
which are evaluated on benchmark datasets.
Examples include 
GlossBERT \cite{huang2019}, 
EWISE \cite{kumar2019}, 
and 
EWISER \cite{bevilacqua2020}.
Because they require labelled training data in the target language,
such systems are generally impractical for languages other than English,
nor are they directly comparable to our proposed methods.

{\bf Knowledge-Based WSD} systems
remain important due to the limited coverage of existing annotated corpora, 
as well as their English bias.
These include graph-based systems such as UKB \cite{agirre2014},
UKB enhanced with SyntagNet \cite{maru2019},
and systems based on
multilingual BERT \cite{devlin2019} and BabelNet \cite{navigli2012},
such as 
SensEmBERT \cite{scarlini2020a}.
We compare our unsupervised results
to both UKB+SyntagNet and SensEmBERT.

{\bf Semi-Supervised Corpus Tagging} systems
depend on sense annotated corpora in one language
to produce sense annotations in other languages.
The current state-of-the-art method in this setting
is MuLaN \cite{barba2020},
which propagates sense annotations from SemCor
and WordNet Gloss Corpus \cite[WNG,][]{langone2004}
to semantically similar contexts in Wikipedia corpora
using contextual word representations from mBERT.
Our {\semisup} method
differs by
leveraging machine translation
to directly
propagate sense annotations across word alignment links.

{\bf Unsupervised Corpus Tagging} systems
produce sense annotations ``from scratch''.
Train-O-Matic \cite{pasini2017}
annotates Wikipedia in multiple languages
by applying the Personalized PageRank (PPR) algorithm to BabelNet.
OneSeC \cite{scarlini2019} combines Wikipedia categories
and BabelNet synset representations to produce WSD training data,
and outperforms Train-O-Matic.
However, both Train-O-Matic and OneSeC annotate nominal instances only,
and hence are not applicable to all-words WSD.
On the other hand, EuroSense \cite{dellibovi2017}
jointly disambiguates content words of all parts of speech
in a parallel corpus using a knowledge-based WSD system.
Our {\unsup} and \labelgen{} methods 
differ in that they explicitly leverage lexical translation information
obtained from a bitext.

{\bf Other work on using translations for WSD:} 
\newcite{resnikyarowsky1999} propose 
to distinguish senses only if a ``minimum subset'' of languages 
translate them differently.
\newcite{apidianaki2009} demonstrates how senses can be induced by
clustering lexical translations,
and proposes an unsupervised WSD system based on
such induced sense inventory and translation information.
\newcite{lefever2011}
frame WSD as translation selection,
and propose a method based on multilingual feature vectors.
Finally, \newcite{taghipour2015} 
annotate English words with their Chinese translations
using manually crafted sense-to-translation mappings.
These methods are not comparable with our work as 
they do not link their sense annotations to the WordNet sense inventory,
and therefore are not applicable to modern WSD datasets.

\section{Semi-Supervised {\semisup}}
\label{sec:method}

\begin{figure}[t]
  \centering
  \includegraphics[keepaspectratio, width=\columnwidth]
  {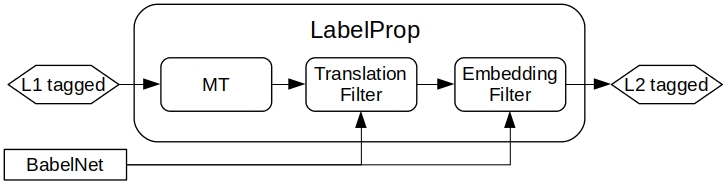}
 \caption{\semisup{} propagates senses from language L1 to language L2.}
 \label{fig_semisup}
\end{figure}

In this section, we introduce {\semisup}, a novel
label propagation approach 
for constructing multilingual sense-annotated corpora.
The idea is to 
translate a sense-annotated corpus 
in order to propagate the sense tags
across the translations.
No sense-annotated data is required in the target language.
The method is composed of three steps:
translation identification, knowledge-base filtering, 
and nearest neighbor filtering
(Figure~\ref{fig_semisup}).

\subsection{Translation Identification}
\label{sec:trans}

Given a sense-annotated source corpus,
we first translate the corpus into the target language
using pre-trained neural machine translation models.
Each sentence 
containing at least one source sense-annotated word
is translated independently.
If the translation of an annotated source word 
can be identified through word alignment, 
we annotate the translation with the same BabelNet synset
as the aligned source word.
This procedure is
based on the assumption that 
lexical translations in context are semantically equivalent,
and therefore very likely to
express the same concept \cite{hauer2020set}.

For alignment, we use \newalign{} \cite{luan2020},
a high-precision alignment tool
which leverages translation information from BabelNet
to improve on a base alignment system. 
In particular, \newalign{} augments the input corpus 
with lexical translation pairs
to bias the aligner towards aligning words which are mutual translations.
It also corrects alignments to maximize the number of aligned words
that share BabelNet synsets.
This emphasis on recovering word-level translation information
makes \newalign{} particularly well-suited to our method.

\subsection{Knowledge-Based Filtering}
\label{sec:filter1}

The sense-projection procedure in the previous step 
may annotate a word with a BabelNet synset
which does not actually contain that word.
These invalid sense annotations may occur
due to non-literal translation
(i.e., the word and its translation do not express the same concept),
errors in translation or alignment,
or omissions in BabelNet.
Since each sense of a word must correspond to a specific synset, 
such invalid annotations are discarded.

\subsection{Nearest Neighbor Filtering}
\label{sec:filter2}

In order to further increase the precision,
we apply a semi-supervised WSD method
to each target translation that is sense annotated
by the previous steps.
For each word,
we verify that the annotation propagated from the source-language corpus
matches the annotation assigned by the WSD system;
otherwise,
we discard that sense annotation.

Our semi-supervised WSD method uses a one-nearest-neighbor approach
with ARES multilingual synset embeddings
\cite{scarlini2020b}.
We first obtain contextual word representations 
of each sense-annotated target translation 
by taking the sum of the last four layers of 
multilingual BERT
\cite{devlin2019}.
Since ARES embeddings have twice the size of the original mBERT embeddings, 
we concatenate each obtained word representation with itself. 
We then compute the cosine similarity between 
the mBERT representation of the word,
and the ARES representation of each synset containing the word.
The synset that maximizes the similarity is taken as
the output of this WSD system.
To reiterate, we retain only the sense annotations from the previous step
that agree with this WSD system.

\section{Unsupervised Symmetric {\unsup}}
\label{sec:unsupmethod}

The \semisup{} method, presented in Section \ref{sec:method}, 
is able to leverage existing sense annotated corpora, such as SemCor,
to create comparable sense annotated corpora in other languages.
However, 
the availability of sense-annotated corpora
in other domains and languages is very limited.
On the other hand, large bitexts are relatively easy to obtain
for many language pairs and domains.

\begin{figure}[t]
  \centering
  \includegraphics[keepaspectratio, width=\columnwidth]
  {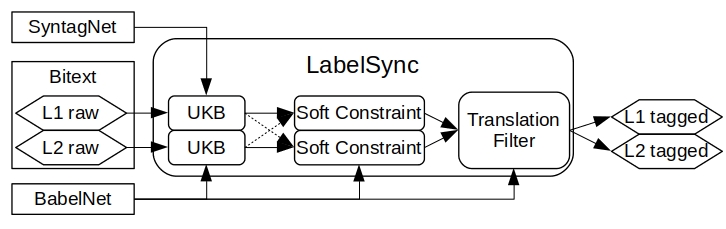}
 \caption{\unsup{} assigns and refines sense annotations in 
two languages simultaneously.}
 \label{fig_unsup}
\end{figure}

To further reduce the dependency of WSD systems
on \emph{any} pre-existing annotated data,
we introduce \unsup{},
a method which annotates both sides of a given bitext.
This method retains the idea of using word alignment
to validate sense annotations,
while eschewing the need for a sense annotated corpus.
It is composed of three steps:
monolingual word sense disambiguation,
multilingual post-processing,
and translation-based filtering
(Figure \ref{fig_unsup}).

\subsection{Monolingual WSD}
\label{sec:ukb}

Our goal is to enrich both sides of the input bitext
with sense tags.
Since \unsup{} does not assume access to any sense-annotated corpus,
we employ
a language-independent knowledge-based WSD system:
a variant of UKB enhanced with SyntagNet \cite{maru2019}.
After
each side of the bitext is annotated independently,
we have two sense annotated corpora,
one in each of the languages represented in the bitext.

\subsection{Multi-Lingual Post-Processing}
\label{sec:twsd}

Now that both sides of the bitext are annotated independently,
we leverage the lexical translation information
inherent in the bitext
to increase the accuracy of the sense annotations.
To improve the performance of our base WSD system,
we employ
the \softconstraint{} method
of \newcite{luan2020}.
This method is applicable to any base WSD system which assigns
a numerical score, such as a probability, 
to each sense of a disambiguated word.
Most modern WSD systems, including UKB, satisfy this property.

The \softconstraint{} method depends on word-level translations
of each annotated word, as well as translation information from BabelNet,
which is based on the hypothesis that the translation of a word token
provides semantic information about its sense \cite{hauer2020set}.
In our case,
translation information is readily available from the bitext.
As with our \semisup{} method,
we use \newalign{} \cite{luan2020}
to word align the bitext.
For each sense-annotated token, 
the aligned word or phrase 
is treated as its translation.

The \softconstraint{} method can also incorporate
sense frequency information,
to bias the annotations toward
more probable senses.
However,  
we exclude sense frequency information from this step,
as it provided no discernible benefit in our development experiments.

\subsection{Translation-Based Filtering}
\label{sec:kbfiltering}

In the final step,
we aim to further reduce the noise in our sense-annotated corpora
by employing a BabelNet-based filtering method, 
similar to the one described in Section \ref{sec:filter1}.
As before, the key idea is to impose 
two constraints on our sense annotations: 
(1) a word should only be annotated with a synset 
that contains the word,
and
(2) aligned words should be annotated with the same synset.
\semisup{} initially guarantees only the latter constraint,
so it has to discard some annotations to ensure the former.
In contrast, \unsup{} initially guarantees the first constraint,
as UKB can only annotate a word with a synset containing it.
However, since each side of the corpus is annotated independently,
the second constraint may not hold.
The final step of {\unsup} is aimed at resolving this problem by
synchronizing the sense annotations across both sides of the bitext.

Unlike
\newcite{dellibovi2017},
who leverage embeddings of concepts to filter questionable annotations,
we adopt a binary alignment-based criteria
using the assumption of semantic equivalence of lexical translations.
We retain only those annotations
that refer to the same multilingual synset as
the sense annotations of their translations.
We also retain annotations if the token cannot be aligned,
or if its translation is not annotated.

\section{Unsupervised Asymmetric {\labelgen}}
\label{labelgen}

Unlike \unsup{}, 
our second unsupervised method, \labelgen{}, 
assumes that the source language is English,
and treats the two sides of the bitext differently.
The goal
is to leverage available English resources
to improve WSD performance on other languages,
rather than 
to generate English sense annotations.

\begin{table*}[t]
\begin{center}
\begin{small}
\setlength{\tabcolsep}{2pt}
\begin{tabular}{lcccc|ccc|cccccc}
   &\multicolumn{4}{c}{\semisup{}}  &\multicolumn{3}{c}{\unsup{}} & \multicolumn{3}{c}{\labelgen{}} \\ 
\hline
        & \thead{Annotated \\ Tokens} & \thead{Annotated \\Word Types} & \thead{Sense\\ Types} & \thead{Failed\\ Alignments} & \thead{Annotated \\ Tokens} & \thead{Annotated \\Word Types} & \thead{Sense\\ Types} & \thead{Annotated \\ Tokens} & \thead{Annotated \\Word Types} & \thead{Sense\\ Types} \\ \hline
EN & -         &  -          &  -          &    -       & 1,783,334  & 9,509 & 16,748 & -  & - & - \\
IT & 399,569   &  25,361     &  29,290     &    30,763  & 2,083,741  &  10,910     &  22,211  & 1,372,876  & 8,355 & 16,046  \\
ES & 403,797   &  25,874     &  31,420     &    31,640  & 1,692,232  &  10,549     &  25,181   & 1,326,244  & 7,926 & 17,335 \\
FR & 407,590   &  25,193     &  32,129     &    32,181  & 1,458,588  &  7,776     &  11,529   & 1,433,647  & 7,712 & 17,980    \\
DE & 309,926   &  23,786     &  23,433     &    64,085  & 645,289    &  2,139     &  2,756     & 821,552  & 6,589 & 9,121    \\ \hline
\end{tabular}
\end{small}
\end{center}
\caption{Statistics of 
the sense-annotated corpora produced by each of our methods.}
\label{tab:semisup-stats}
\end{table*}

\subsection{English WSD}
\label{labelgen1}

Given a bitext,
we first apply a knowledge-based WSD system
to the English side \emph{only},
as described in Section \ref{sec:ukb}.
The lexical information for other languages 
is retrieved from BabelNet multi-synsets
which are aligned to WordNet 3.0 synsets. 
This automatic candidate retrieval process is noisy,
because most BabelNet lexicalizations are automatically generated
from various resources. 
In addition, 
while English WordNet contains the sense frequency estimates 
from the manually-annotated SemCor,
such information is not readily available for other languages.
Hence, WSD annotations are more accurate
for English compared to other languages.

\subsection{Label Propagation}
\label{labelgen2}

Having automatically sense-tagged the English side of the bitext,
we propagate the labels to the non-English side
using the procedure described in Section \ref{sec:trans}.
In effect, we are applying the first part of \semisup{},
treating the English side as a sense-tagged corpus,
and the other side as its translation.
At the end of this process, 
both sides of the bitext are sense-annotated.

\subsection{Re-Ranking and Filtering}
\label{labelgen3}

We further refine the sense annotations on
the non-English side of the bitext.
We first apply the \softconstraint{} method
as described in Section \ref{sec:twsd},
which re-ranks the possible senses for each annotated word
using the assigned WSD scores.
We then apply the filtering procedure from Section \ref{sec:filter1},
which removes any sense annotations
that do not exist in the BabelNet sense inventory.

\section{Evaluation}
\label{sec:exp}

Following prior work,
we extrinsically evaluate our corpus construction approaches 
by providing the generated annotations 
as training data for supervised WSD systems
({\em reference systems}),
which are then evaluated on standard multi-lingual WSD benchmarks.
While our methods could also be applied to low-resource languages,
the current lack of evaluation datasets 
precludes such experiments in this work.

\subsection{Reference Supervised WSD Systems}
\label{refs}
 
We perform experiments with
two reference supervised WSD systems:
(1) IMS
\cite{zhong2010}
with the most-frequent-sense (MFS) backoff for English,
and (2) \emph{mBERT},
a transformer-based method,
built on multilingual BERT \cite{devlin2019},
as described by \newcite{barba2020}.
We use the default parameter settings and number of training 
epochs\footnote{https://github.com/edobobo/transformers-wsd}.
We train each model on each
set of automatically produced sense annotations.

Following prior work,
we use the SemEval-2007 dataset \cite{raganato2017}
as our validation set for the English experiments.
Because of the lack of standard validation sets for non-English languages,
we use random samples of 1000 sentences from our training corpora.
The hyperparameters of each system 
are held constant throughout all experiments.

\subsection{Test Data}
\label{testsets}

We test the reference WSD models on 
standard multilingual benchmark datasets:
SemEval-2013 task 12 \cite{navigli2013},
which contains data for Italian, Spanish, French, and German,
and SemEval-2015 task 13 \cite{moro2015},
which covers Italian and Spanish.
The SemEval-2013 datasets contain only nominal instances,
while the SemEval-2015 datasets cover nouns, verbs, adjectives, and adverbs.
We use the latest version of the 
datasets\footnote{https://github.com/SapienzaNLP/mwsd-datasets},
which are annotated with synsets from BabelNet version 4.0.

For the experiments on English (Section \ref{english}),
we use the standardized benchmarks of
\newcite{raganato2017}\footnote{http://nlp.uniroma1.it/wsdeval},
which comprise all-words test sets from five shared tasks:
Senseval2 \cite{edmonds2001}, 
Senseval3 \cite{snyder2004}, 
SemEval-2007 \cite{pradhan2007}, 
SemEval-2013 \cite{navigli2013}, and 
SemEval-2015 \cite{moro2015}.
We also report the average results 
on the concatenation of all five test sets,
which we refer to as ALL.

\subsection{Semi-Supervised Approaches}
\label{eval_semi}

This section is devoted to 
the empirical evaluation of the \semisup{} method
from Section~\ref{sec:method}.

\subsubsection{Experimental Setup}

We apply {\semisup}
to a sense-annotated English corpus
comprised of
SemCor \cite{miller1994} 
and the WordNet Gloss Corpus (WNG) \cite{langone2004}.
Following \newcite{luan2020},
we translate each sentence of our English corpus
with Google Translate
independently into Italian, Spanish, French, and German.
As described in Section \ref{sec:trans},
we induce word alignment by applying
\newalign{}
with \oldalign{} \cite{dyer2013} as its base aligner.
Table \ref{tab:semisup-stats} 
contains the statistics for the corpora created
with our methods.
``Sense types'' indicates the number of distinct word senses in the corpus.
``Failed alignments'' refers to the number of
English sense annotations that could not be propagated.

We compare \semisup{} to MuLaN \cite{barba2020},
the current state-of-the-art system for semi-supervised corpus annotation,
which also uses SemCor+WNG as its manually-annotated base English corpus.
Specifically,
we apply the same procedure 
to train the supervised reference system (IMS or mBERT)
on the annotated data produced by each method.
We train a single model for each system,
using only the corpus produced by that system for each language,
which limits the impact of language-specific issues.
Nevertheless,
due to both software and hardware variables and hyper-parameters, 
our MuLaN results differ from those reported in the original paper.

When reporting the results achieved with mBERT,
we also include the results of two recent WSD systems:
ARES, using the reported results from \newcite{scarlini2020b},
and 0-Shot WSD, with results replicated using
the code provided by \newcite{barba2020}.
Since they are not designed to create annotated training data for
other WSD systems, they are not directly comparable to {\semisup}.

\subsubsection{Results}
\label{sec:semisupres}

\begin{table}[t]
\begin{center}
\begin{small}
\setlength{\tabcolsep}{2pt}
\begin{tabular}{lcccccccc}
\hline
 & \multicolumn{4}{c}{SemEval-2013}        & & \multicolumn{2}{c}{SemEval-2015} &          \\  \cline{2-5} \cline{7-8}
        Model   & IT  & ES  & FR & DE & & IT & ES & AVG  \\ \hline
        MCS     & 44.2   & 37.1   & 53.2  & {\bf 70.2}    & & 44.6   & 39.6 & 48.2 \\ 
	MuLaN          & 65.6   & 65.6   & \textrm{68.1}  & \textrm{69.7}  & & 63.7   & 59.9 & 65.4 \\ 
	\semisup{}    & \textrm{71.4}  & \textrm{71.0}  & 65.1   & 62.8  & & \textrm{67.1}  & \textrm{64.0}  &   \textrm{66.9} \\ 
\hline
-NN				& 70.5		& 70.1		& 62.7 & 63.5		& & 66.3		& 61.8 	& 65.8 	\\
-NN -KB		& 66.7		& 68.2		& 60.2		& 56.7		& & 61.2   	& 60.2	 	& 62.2 	\\ 
\hline
\end{tabular}
\end{small}
\end{center}
\caption{WSD F-score obtained with IMS trained on 
the corpora generated by MuLaN and \semisup{}.}
\label{tab:semisup-ims}
\end{table}

\begin{table}[t]
\begin{center}
\begin{small}
\setlength{\tabcolsep}{2pt}
\begin{tabular}{lcccccccc}
\hline
	& \multicolumn{4}{c}{SemEval-2013}				& & \multicolumn{2}{c}{SemEval-2015} & 		\\ \cline{2-5} \cline{7-8} 
Model					& IT				& ES				& FR 				& DE				& & IT 				& ES				& AVG			\\ \hline
ARES					& 77.0 & 75.3				& \textrm{81.2} & \textrm{79.6}				& & 71.4 & \textrm{70.1}				& 75.7				\\ 
0-shot\textsubscript{SC+WNG}
			& 78.3 & 77.6 & 80.8 & 78.3 && 70.5 & 68.6 & 75.7 \\ 

\hline

MuLaN				& 76.8				& 78.4				& 80.4	& 78.8	& & 68.7				& 67.8				& 75.2				\\ 
	\semisup{}					& \textrm{78.4}	& \textrm{78.5}  & 80.4			 	& 77.8 & & \textrm{72.7}	& 68.9	& \textrm{76.1}	\\ 
\hline
\end{tabular}
\end{small}
\end{center}
\caption{WSD F-score obtained with mBERT trained on 
the corpora generated by MuLaN and \semisup{}.
Results of two semi-supervised WSD systems
are included for reference.}
\label{tab:main}
\end{table}

Table \ref{tab:semisup-ims}
presents the multilingual WSD results 
obtained by IMS on standard test sets.
While IMS
is no longer a state-of-the-art system,
it is still commonly used as a benchmark 
for evaluating automatically generated corpora \cite{scarlini2019}.
The results demonstrate the relative quality of the generated corpora:
\semisup{} is better than MuLaN 
on Italian and Spanish, as well as on average.
The difference in performance is found 
to be statistically significant across all six datasets
($p < 0.05$ with McNemar's test). 
Neither of the approaches outperforms the 
most common sense (MCS) baseline on German,
which we discuss further
in Section \ref{semisup_error_analysis}.
The ablation results in the last two rows show that
both the nearest neighbour WSD filter (NN)
and the translation-based filter (KB)
improve the quality of the annotations.

Table \ref{tab:main} presents the corresponding results
using the more recent mBERT as the reference system.
Our results are slightly better on average than those of 0-shot and ARES.
However, only the MuLaN results are directly comparable 
to our \semisup{} results,
as both systems produce training data for a supervised WSD reference system.
{\semisup} matches or outperforms MuLaN 
on every dataset except German SemEval-2013,
and achieves better results on average
compared to the results we replicated.
The difference in F-score between \semisup{} and 
our replicated MuLaN experiment
is significant for the SemEval 2015 Italian dataset
($p < 0.05$ with McNemar's test).

\subsubsection{Error Analysis}
\label{semisup_error_analysis}

Error analysis suggests two reasons for the relatively 
low results on the German data.
First, 
English multi-word compounds often correspond to single words in German,
which makes it difficult to properly propagate English sense annotations.
For example, 
the two words in \textit{\textrm{giveaway program}},
which is a translation of \textit{\text{Werbeprogramm}},
are separately annotated with different senses. 
The second issue
is the quality of the BabelNet translation coverage. 
We observe that 
among 69,402 BabelNet synsets, 
that correspond to word senses appearing in SemCor+WNG, 
only 40,490 synsets contain at least one German translation,
compared to over 50,000 synsets in
each of the other three languages.

\subsection{Unsupervised Approaches}
\label{eval_unsup}

In this section,
we evaluate  
our unsupervised methods, 
\unsup{} and \labelgen{},
against comparable systems.

\subsubsection{Experimental Setup}
\label{experimental_setup}

We adopt UKB \cite{agirre2014}
as the base knowledge-based WSD system used in the first step of 
both \unsup{} and \labelgen{}
to perform the initial tagging of a bitext.
(This base WSD system is not to be confused with
the reference supervised WSD system that is only used for the purpose
of corpus  evaluation.)
Following \newcite{maru2019},
we use WordNet as a lexical knowledge base,
enriching it with information from WNG,
and syntagmatic information from SyntagNet.
BabelNet is the source of multilingual lexicalization information.
When applying UKB,
the PPR\textsubscript{w2w} variant of the personalized PageRank algorithm
is run separately for each word, 
while concentrating the initial probability mass
in the senses of the context words rather than the focus word.

Both of our unsupervised methods operate on  
an unannotated bitext.
To keep the corpus size manageable,
we randomly sample 200k sentences
with English, French, German, Italian, and Spanish translations
from EuroSense \cite{dellibovi2017}
discarding its existing sense annotations.
This produces four bitexts
with English as one of the languages,
which we align at the word level using \mbox{\newalign{}}.
The \softconstraint{} method employed by \unsup{}
to refine the initial sense annotations 
leverages the lexical translations.
Table~\ref{tab:semisup-stats} presents the statistics 
of the produced corpora.

The direct competitor
of \unsup{} and \labelgen{}
is OneSeC \cite{scarlini2019},
an unsupervised system which produces sense-annotated data
by leveraging the semantic information within Wikipedia categories.
Since OneSeC can only tag nouns,
any model trained on a corpus it produces
will likewise only be able to disambiguate nouns.
Therefore,
we do not apply models trained on OneSeC to the SemEval-2015 datasets,
which include verb, adjective, and adverb instances.
For our multilingual experiments,
we also compare to two knowledge-based WSD systems
described in Section~\ref{sec:rw}:
UKB with SyntagNet \cite{maru2019}, 
and {\sensembert{} \cite{scarlini2020a}}.

\subsubsection{Multilingual Results}

\begin{table}[t]
\begin{center}
\begin{small}
\setlength{\tabcolsep}{2pt}

\begin{tabular}{lcccccccc}
\hline
  & \multicolumn{4}{c}{SE-2013}        & & \multicolumn{2}{c}{SE-2015} &          \\  \cline{2-5} \cline{7-8}
  Model   & IT  & ES  & FR & DE & & IT & ES & AVG   \\ \hline
  MCS     & 44.2   & 37.1   & 53.2  & 70.2    & & 44.6   & 39.6 & 48.2   \\
  UKB+SyntagNet   & 72.1   & 74.1   & 70.3  & 76.4  & & 69.0 & 63.4 & 70.9 \\
  \sensembert{}   & 69.8   & 73.4   & 77.8  & \textrm{79.2}  & & -    & -    & -   \\ \hline
 
  OneSeC          & 63.5   & 61.6   & 65.1  & 75.8  & & -    & -    & -   \\
  \unsup{}    & 75.7  & 78.2  & 72.4   & 75.3  & & \textrm{70.8}  &
\textrm{66.3}  & 73.1   \\
  \labelgen{}    & \textrm{77.8}  & \textrm{80.5}  & \textrm{80.7}  &
75.4  & & 68.7  & 66.1  & \textrm{74.9}   \\       
\hline
\end{tabular}
\end{small}
\end{center}
\caption{WSD F-score obtained with mBERT trained on the corpora generated by
\unsup{} and \labelgen{}.
}
\label{tab:unsuptable}
\end{table}

Table \ref{tab:unsuptable} presents the multilingual WSD results 
when using mBERT as the reference WSD system.
With the consistent exception of German,
the results of mBERT trained on the annotations produced by {\unsup}
are substantially better than those trained on the corpus generated by OneSeC,
which is the previous state-of-the-art for unsupervised corpora tagging.
Unlike OneSeC, our unsupervised methods can annotate 
tokens representing all parts of speech,
and can therefore be applied to the SemEval 2015 datasets.
\unsup{} also outperforms 
both knowledge-based WSD systems,
UKB+SyntagNet and \textsc{SensEmBERT},
and the most common sense (MCS) baseline.
\labelgen{} further improves on
\unsup{}
by 
1.8\% on average.
This makes it our best performing system,
which sets a new state-of-the-art on the SemEval-2013
Italian, Spanish, and French datasets.

\subsubsection{English Results}
\label{english}

\begin{table}[t]
\begin{center}
\begin{small}
\setlength{\tabcolsep}{2pt}
\begin{tabular}{lccccccc}
\hline
	& SE2 & SE3 & S07 & S13 & S15 & ALL	& \\ \hline
MFS & 66.8 & 66.2 & 55.2 & 63.0 & 67.8 & 65.2 \\
SemCor & \textrm{71.3} & \textrm{69.1} & \textrm{61.5} & 65.5 & 68.3 & 68.3 \\
EuroSense + SemCor
& - & - & - & 66.4 & 69.5 & - \\
\hline
\unsup{}
& 69.4 & 64.5 & 57.4 & \textrm{71.7} &
\textrm{72.9} & \textrm{68.4} \\ \hline
\end{tabular}
\end{small}
\end{center}
\caption{WSD F-score on all instances 
obtained with IMS
trained on the corpora generated by \unsup.}
\label{tab:unsuptable-ims}
\end{table}

\begin{table}[t]
\begin{center}
\begin{small}
\setlength{\tabcolsep}{2pt}
\begin{tabular}{lccccccc}
\hline
	& SE2 & SE3 & S07 & S13 & S15 & ALL	& \\ \hline
MFS  &  66.8  &  66.2  &  55.2  &  63.0  &  67.8  &  65.2 \\
SemCor  &  74.8  &  73.1  &  64.2  &  69.9  &  74.7  &  72.6 \\
\hline
\unsup{}  &  69.6  &  65.9  &  55.2  &  71.4  &  75.1  &  68.9 \\
\hline
\end{tabular}
\end{small}
\end{center}
\caption{WSD F-score on all instances 
obtained with mBERT
trained on the corpora generated by \unsup.}
\label{tab:unsuptable-mbert}
\end{table}

In this section, we evaluate \unsup{} on English WSD.
We do not test \labelgen{} on English,
as it was specifically designed to tag non-English corpora.
Furthermore,
because OneSeC annotates only nominal instances,
we conduct separate all-words and nouns-only experiments.

Table~\ref{tab:unsuptable-ims}
presents the English WSD results
on all test instances
with IMS
as the reference system.
The corpus annotated by \unsup{} is 
a subset of the corpus annotated by EuroSense.
The results of {\unsup} on S13 and S15 are much better 
than the results using EuroSense augmented with SemCor
as reported by \newcite{dellibovi2017},
which
we attribute to the explicit use of translation information.
On the concatenation of all five test sets,
the unsupervised IMS+\unsup{} results
rival the supervised results of IMS trained
on SemCor, 
a manually sense-annotated corpus.

In Table \ref{tab:unsuptable-mbert}, we see
mBERT performing much better than IMS when trained on SemCor.
Remarkably,
the corpus generated in an unsupervised manner by \unsup{}
yields results on S13 and S15
that surpass 
those obtained by training mBERT
directly on SemCor.
These results 
are impressive because
\unsup{} makes no use of manual sense annotation.
We speculate that this may be due to the difference in domain
between SemCor and the corpus annotated by \unsup{}.

Tables \ref{tab:english-nouns-ims} 
and \ref{tab:english-nouns-mbert} present
the results of our final set of experiments,
in which
the reference systems 
are tested on English nominal instances only.
Here, we can compare \unsup{} directly to its competitor, OneSeC. 
Both of our reference WSD systems, IMS and mBERT,
clearly perform better across all datasets
when trained on the corpus produced by our \unsup{} method,
compared to training on the corpus produced by OneSeC.
These results establish \unsup{} as the new state of the art
for unsupervised English corpus sense tagging,
and a step towards overcoming the knowledge acquisition bottleneck in WSD.

\begin{table}[t]
\begin{center}
\begin{small}
\setlength{\tabcolsep}{3pt}
\begin{tabular}{lccccccc}
\hline
	& SE2 & SE3 & S07 & S13 & S15 & ALL	& \\ \hline
MFS & 72.1 & 72.0 & 65.4 & 63.0 & 66.3 & 67.6 \\
SemCor & \textrm{76.8} & \textrm{73.8} & 67.3 & 65.5 & 66.1 & 70.4 \\
\hline
OneSeC & 73.2 & 68.2 & 63.5 & 66.5 & 70.8 & 69.0\\
\unsup{} & \textrm{76.1} & \textrm{70.0} & \textrm{68.6} &
\textrm{71.7} & \textrm{72.1} & \textrm{72.3} \\ 
\hline
\end{tabular}
\end{small}
\end{center}
\caption{
English WSD F-score on \underline{nominal instances}
obtained with IMS
as the reference WSD system,
}
\label{tab:english-nouns-ims}
\end{table}

\begin{table}[t]
\begin{center}
\begin{small}
\setlength{\tabcolsep}{3pt}
\begin{tabular}{lccccccc}
\hline
	       &   SE2  &   SE3  &   S07  &   S13  &   S15  &   ALL & \\ \hline
MFS        &  72.1  &  72.0  &  65.4  &  63.0  &  66.3  &  67.6 \\
SemCor     &  \textrm{79.7}  &  \textrm{75.4}  &  \textrm{67.9}  &
69.9  &  75.0  &  \textrm{74.0} \\
\hline
OneSeC     &  74.2  &  67.1  &  62.9  &  68.8  &  74.2  &  70.2 \\
\unsup{}
&  76.8  &  70.8  &  66.0  &
\textrm{71.4}  &  \textrm{75.7}  &  73.0 \\
\hline
\end{tabular}
\end{small}
\end{center}
\caption{
English WSD F-score on \underline{nominal instances}
obtained with mBERT 
as the reference WSD system.
}
\label{tab:english-nouns-mbert}
\end{table}

\section{Conclusion}
\label{sec:con}

We have introduced new methods to address the 
knowledge acquisition bottleneck in word sense disambiguation
in both the semi-supervised and unsupervised settings.
The methods leverage
recent advances in machine translation, alignment, and contextual embeddings.
Extrinsic experiments with a variety of WSD systems demonstrate that 
the quality of 
the corpora created by our methods 
is substantially higher compared to those produced by prior work.
Our methods for automatic sense tagging
can produce annotated corpora for many languages, 
and approach the quality of manual annotation in some cases.
We make our corpora available for further research.

One advantage of our unsupervised methods is that
they can be applied to annotate any bitext involving any languages.
We posit that our results could be further improved
by annotating corpora with broader domain coverage,
or by matching the domain of the source corpus
to the domain of the data to be disambiguated.
We leave this as a direction for future work.

\section*{Acknowledgments}

The authors of this paper are listed in alphabetical order.
Yixing Luan and Arnob Mallik conducted the experiments 
with the semi-supervised and unsupervised methods, respectively.
Bradley Hauer and Grzegorz Kondrak
prepared the final version of the paper.

This research was supported by
the Natural Sciences and Engineering Research Council of Canada (NSERC),
and the Alberta Machine Intelligence Institute (Amii).

\bibliographystyle{acl_natbib}
\bibliography{wsd}

\end{document}